\documentclass{article}

    \usepackage[preprint, nonatbib]{tackling_climate_workshop_style}

\usepackage[utf8]{inputenc} 
\usepackage[T1]{fontenc}    
\usepackage{hyperref}       
\usepackage{url}            
\usepackage{booktabs}       
\usepackage{amsfonts}       
\usepackage{nicefrac}       
\usepackage{microtype}      
\usepackage{graphicx}
\usepackage{subcaption}
\usepackage{wrapfig}
\usepackage{bm}
\usepackage[symbol]{footmisc}
\usepackage{float}
\restylefloat{table}

\title{Cross Modal Distillation for Flood Extent Mapping}

\author{%
  Shubhika Garg \\ Google Research \And
   Ben Feinstein \\ Google Research
    \And
   Shahar  Timnat\\
   Google Research
   \And
   Vishal Batchu \\
   Google Research
   \And
   Gideon Dror \\
   Google Research
   \And
   Adi Gerzi Rosenthal \\
   Google Research
   \And
   Varun Gulshan \\
   Google Research
}

\begin{document}

\maketitle

\begin{abstract}
The increasing intensity and frequency of floods is one of the many consequences of our changing climate. In this work, we explore ML techniques that improve the flood detection module of an operational early flood warning system. Our method exploits an unlabelled dataset of paired multi-spectral and Synthetic Aperture Radar (SAR) imagery to reduce the labeling requirements of a purely supervised learning method. Prior works have used unlabelled data by creating weak labels out of them. However, from our experiments we noticed that such a model still ends up learning the label mistakes in those weak labels. Motivated by knowledge distillation and semi supervised learning, we explore the use of a teacher to train a student with the help of a small hand labelled dataset and a large unlabelled dataset. Unlike the conventional self distillation setup, we propose a cross modal distillation framework that transfers supervision from a teacher trained on richer modality (multi-spectral images) to a student model trained on SAR imagery. The trained models are then tested on the Sen1Floods11 dataset. Our model outperforms the Sen1Floods11 baseline model trained on the weak labeled SAR imagery by an absolute margin of $6.53\%$ Intersection-over-Union (IoU) on the test split.

\end{abstract}

\section{Introduction}

Floods are one of the major natural disasters, exacerbated by climate change, affecting between 85 million to 250 million people annually and causing between $\$ 32$ to $\$36$ billion in economic damages~\cite{guha2015dat,emerton2016continental}. Some of these harms can be alleviated by providing early flood warnings, so that people can take proactive measures such as planned evacuation, move assets such as food and cattle and use sandbags for protection. One of the important user experience elements for an effective warning system is its overall accuracy, as false alerts lead to eroded trust in the system. Our work contributes towards improving the accuracy of flood warning systems such as~\cite{nevo2022flood}, by increasing the accuracy of the inundation module. The inundation model in ~\cite{nevo2022flood} learns a mapping between historical river water gauge levels and the corresponding flooded area which can be leveraged to predict future flooding extent based on forecast of the future river water gauge level. The accuracy of these forecasts is directly correlated with the accuracy of the underlying historical segmentation maps, hence we aim to improve the segmentation module through our contributions in this work.

In recent years, remote sensing technology has considerably improved and helped us in the timely detection of floods and monitoring their extent. It provides us with satellite data at different spatial resolutions and temporal frequencies. For example, MODIS ~\cite{justice1998moderate} provides low resolution data(250m) with high temporal frequency(~2days). There are medium spatial resolution satellites, from 10m to 30m range, such as Sentinel-2~\cite{drusch2012sentinel}, Sentinel-1~\cite{torres2012gmes} and Landsat~\cite{roy2014landsat} available, but they have a slightly lower temporal frequency (~6-15 days). High resolution data with resolution ranging from a few centimeters to 1m, can also be obtained on special demand using airborne radars. However, the process of obtaining airborne data is very expensive. Hence, Sentinel-1 and Sentinel-2 satellites are preferred to map the surface of water because they provide a good trade-off between both spatial and temporal resolutions along with open access to their data. Although, Sentinel-2 is better for water segmentation because it shows high water absorption capacity in short wave infrared spectral range (SWIR) and near infrared (NIR) spectrum, it cannot penetrate cloud cover. This limits its application for mapping historical floods as cloud cover is highly correlated with flooding events. On the other hand, radar pulses can readily penetrate clouds, making SAR data provided by Sentinel-1 satellite well suited for flood mapping~\cite{tarpanelli2022effectiveness,mason2021floodwater,vanama2020gee4flood}. 

Thresholding algorithms~\cite{liang2020local,martinis2015backscatter,brown2016progress} are traditionally used to segment flooded regions from SAR images since water has a low back scatter intensity. Commonly used techniques like Otsu thresholding ~\cite{bao2021water} assume that the histogram of a SAR image has a bimodal distribution and they work well for many cases. However, its failure modes include generating false positives for mountain shadows and generating excessive background noise due to speckle in SAR imagery. This noise can be removed using Lee speckle filters ~\cite{lee1981refined} or mean filters, however this results in small streams being missed out. In recent years, Convolutional Neural Networks (CNN) have been used to segment flooded areas from satellite images. Unlike traditional pixel-wise methods, they can look at a larger context and incorporate spatial features from an image. A lot of work has focused on using opportunistically available cloud free Sentinel-2 images~\cite{mateo2021towards, akiva2021h2o}. Though these methods have good performance, their utility at inference time is limited because of the cloud cover issues mentioned previously. Another line of work fuses Sentinel-1 and Sentinel-2 images~\cite{konapala2021exploring, drakonakis2022ombrianet, tavus2020fusion, bai2021enhancement} to enhance surface water detection during flooded events. These methods not only a require a cloud free Sentinel-2 image, but also require that both images are acquired at about the same time to avoid alignment issues. There also has been some work done that uses multi temporal images~\cite{zhang2020use, sunkara2021memory, yadav2022attentive} containing a pre-flood and a post-flood event. These methods do change detection and exhibit better performance. In our work however, we focus on methods that only take a single Sentinel-1 timestamp image as input.

 Most of the prior work that uses a single Sentinel-1 image as input~\cite{ghosh2022automatic,katiyar2021near,helleis2022sentinel} for flood segmentation, makes use of Sen1Floods11~\cite{bonafilia2020sen1floods11} - a publicly available dataset with a small set of high quality hand labelled images and a larger set of weak labelled images. However, the limitation of using weak labelled data (despite using various regularization techniques), is that the model still learns the mistakes present in those labels. Previously, works such as ~\cite{wang2017multiclass, song2019selfie, huang2020self, zheng2021meta} have been explored to handle noisy label using loss adjustment. \textit{Wang et al.} ~\cite{wang2017multiclass} uses loss re-weighting to assign a lower weight to incorrect labels, ~\cite{song2019selfie, huang2020self} employ label refurbishing by using entropy to correct noisy labels or by doing a progressive refinement of the noisy labels using a combination of the current model output and the label. \textit{Zheng et al.} ~\cite{zheng2021meta} proposes a meta-learning based framework, where a label correction model is trained to correct the noisy labels and the main model is trained on these corrected labels. However, most of these methods have only been explored for classification task and are hard to optimize for segmentation tasks due to increased complexity. Our work also explores methods to do label correction, however we use a simpler idea of using historical temporal imagery to correct weak labels during training as described in Sec \ref{subsec:weak_label_improv}.

In contrast to the above work, we also explore leveraging unlabelled data through semi-supervised learning techniques as opposed to creating weak labels. Methods such as ~\cite{sohn2020fixmatch, wang2022semi, ahmed2022cleaning, paul2021flood} have explored semi supervised techniques to make use of both labelled and unlabelled data. The basic principle of ~\cite{sohn2020fixmatch} is that it uses pseudo labels predicted on the weak augmented image, to consistently train heavily augmented images. \textit{Ahmed et al.} ~\cite{ahmed2022cleaning} ensembles predictions from multiple augmentations to produce more noise resilient labels. However, most of these works focus on RGB images and their performance degrades in remote sensing images. This happens because most of the augmentations are hand crafted for RGB images only.
Motivated to use different modalities of satellite data of the same location as natural augmentations in semi-supervised methods and cross modal feature distillation~\cite{gupta2016cross}, we use a teacher student setup that extracts information from a more informative modality (Sentinel-2) to supervise paired Sentinel-1 SAR images with the help of a small hand labelled and a large unlabelled data set. Similar to~\cite{gupta2016cross}, we transfer supervision between different modalities. However, instead of supervising an intermediate feature layer like~\cite{gupta2016cross}, we transfer supervision at the output layer and apply this towards a new application (i.e. flood segmentation). Our main contribution in this work are:
\begin{itemize}
    \item We propose a cross modal distillation framework and apply it to transfer supervision between different modalities using paired unlabelled data.
    \item We propose a method to improve the quality of weak labels using past temporal data and use that to enhance the weak label baseline.
    \item We curate an additional large dataset (in addition to Sen1Floods11) from various flooding events containing paired Sentinel-1 and Sentinel-2 images and a weak label based on Sentinel-2 data.
\end{itemize}

\section{Data}

\subsection{Data features}
\label{subsection:input}

\textbf{Sentinel-1 image}: Sentinel-1~\cite{torres2012gmes} mission launched by European Space Agency (ESA) consists of 2 polar orbiting satellites to provide free SAR data. However, recently one of the satellite malfunctioned and currently remains out of service. ESA has plans to launch another satellite in 2023. This satellite is an example of active remote sensing satellite and uses radio waves operating at a centre frequency of 5.405 GHz. This allows it to see through cloud cover. It has a spatial resolution of 10m and has a return period of 6 days. We use the Sentinel-1 GRD product and utilize the bands that consist of dual polarized data: Vertical Transmit-Vertical Receive (VV) and Vertical Transmit-Horizontal Receive (VH). These bands representing the backscatter coefficient are converted to logarithmic(dB) scale. The backscatter coefficient is mainly influenced by the physical characteristics such as roughness and the geometry of the terrain and the dielectric constant of the surface. It is discriminative for detecting surface water as water reflects away all the emitted radiation from the satellite in a specular way.

\textbf{Sentinel-2 image}: Sentinel-2~\cite{drusch2012sentinel} mission was also launched by the European Space Agency (ESA) and has 2 polar orbiting satellites. It provides multispectral data at a resolution of 10m and has a return period of 5 days. We use the L1-C Top of Atmosphere (TOA) product. It is a passive remote sensing satellite operating in visible and infrared wavelength. Its images are affected by atmospheric conditions and often contain significant cloud cover. The multispectral data consists of 13 bands and in this work we use 4 bands: B2 (Blue), B3 (Green), B4 (Red) and B8 (NIR). These bands are used for various tasks like land cover/use monitoring, climate change and disaster monitoring. 

\textbf{Weak label}: The weak labels are computed from the Sentinel-2 image. Though cloud free Sentinel-2 data is rarely available during a flooding event, we can still opportunistically sample timestamps from a long time range and get enough cloud-free views. Initially, the cloud mask is estimated using Sentinel-2 quality assurance band (QA60). The cloud mask is then dilated to mask out the nearby cloud shadows as the spectral signature of the cloud shadows is similar to that of water~\cite{li2013automatic}. Weak flood labels are then created by thresholding the Normalized Difference Water Index ($NDWI =(B3-B8)/(B3+B8)$) band. The pixels having value greater than 0 are marked as water and the rest are marked dry. 

\textbf{Water occurrence map}: The water occurrence map~\cite{pekel2016high} shows how water is distributed temporally throughout the 1984-2020 period. It provides the probability of each pixel being classified as water averaged over the above time period. It has a 30m resolution data. The data was generated using optical data from Landsat 5, 7, and 8. This map can be used to capture both the intra and inter-annual changes and to differentiate seasonal/ephemeral flooding pixels from permanent water pixels. Each pixel was individually classified into water / non-water using an expert system and the results were collated into a monthly history for the entire time period.  Averaging the results of all monthly calculations gives the long-term overall surface water occurrence. 

\subsection{Datasets}
\textbf{Sen1Floods11 dataset}: This is is a publicly available dataset~\cite{bonafilia2020sen1floods11}, containing 4831 tiles from 11 flooding events across 6 continents. It contains paired Sentinel-1 SAR and Sentinel-2 multi-spectral images. Each tile is $512{\times}512$ pixels at a resolution of 10m per pixel. Due to the high cost of labeling, only 446 tiles out of 4831 are hand labelled by remote sensing experts to provide high quality flood water labels. The authors provide an IID split of these hand labelled tiles, containing 252 training, 89 validation, and 90 test sample chips. The remaining 4,385 tiles have weak labels prepared by thresholding NDVI (Normalized Difference Vegetation Index) and MNDWI (Modified Normalized Difference Water Index) values. The weak labels are only used for training as they are not accurate enough to be used in validation or test. We also augment the dataset with the water occurrence map for every tile.

\textbf{Floods208 dataset}: We curated additional imagery by downloading closely acquired Sentinel-1 and Sentinel-2 images from Earth Engine \cite{gorelick2017google} during flood events provided to us by external partners. 


The data is extracted using the following steps:
\begin{itemize}
    \item For each data point consisting of latitude, longitude and flooding event timestamp, get the Sentinel-1 image of the Area of Interest (AOI).
    \item Search for overlapping Sentinel-2 images within 12hrs of Sentinel-1 timestamp. Filter out images with $> 12\%$ cloud cover.
    \item Pick the Sentinel-2 image closest to Sentinel-1 timestamp from the filtered images. If none are available, we discard this data point.
\end{itemize}

\begin{figure*}[]
\centering
\includegraphics[width=13cm]{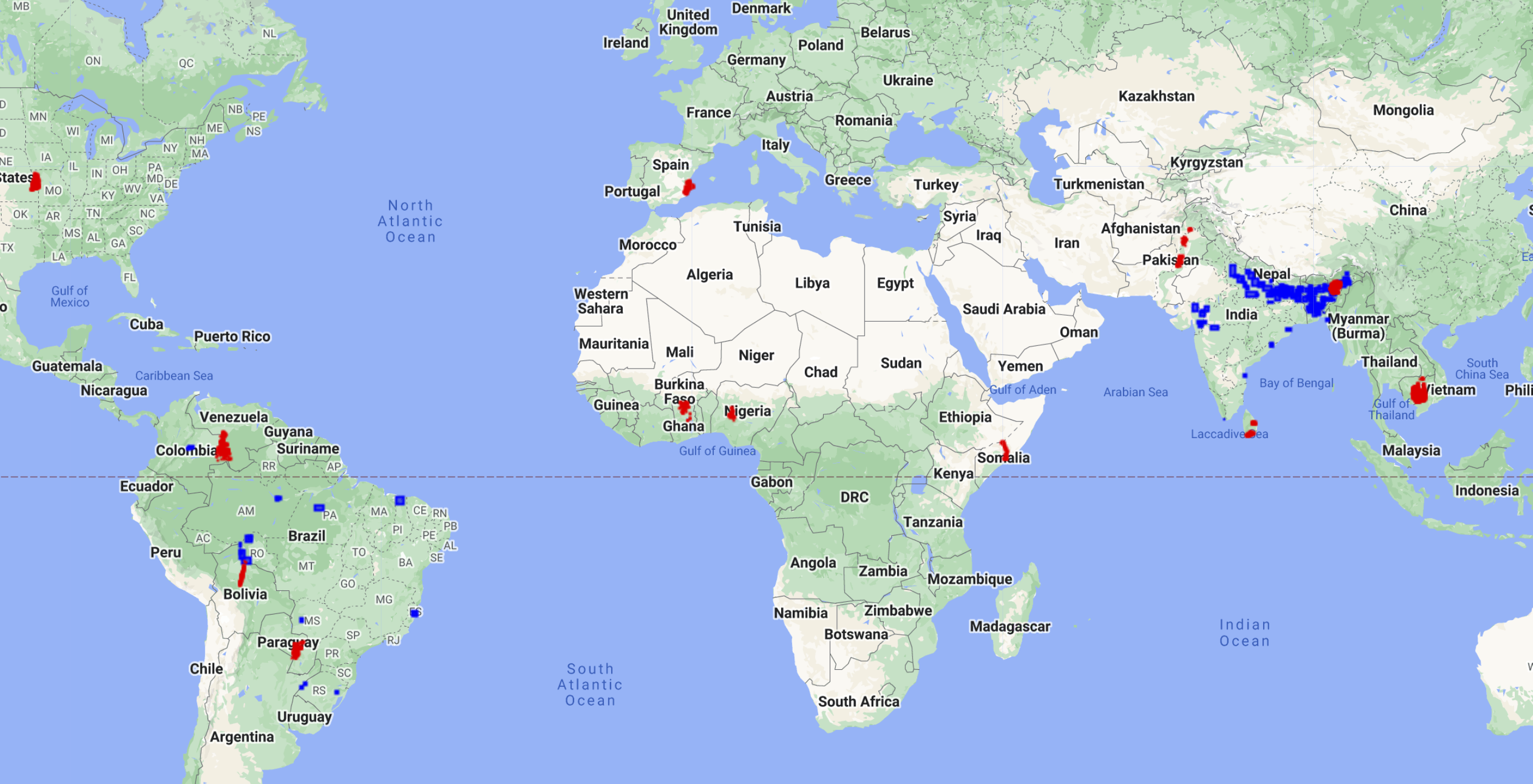}
\caption{Red points highlight the regions from where Sen1Floods11 flooding event data points were sampled and blue points indicate the same for Floods208 dataset.}
\label{fig:s1}
\end{figure*}

\begin{figure*}[]
\centering
\includegraphics[width=13cm]{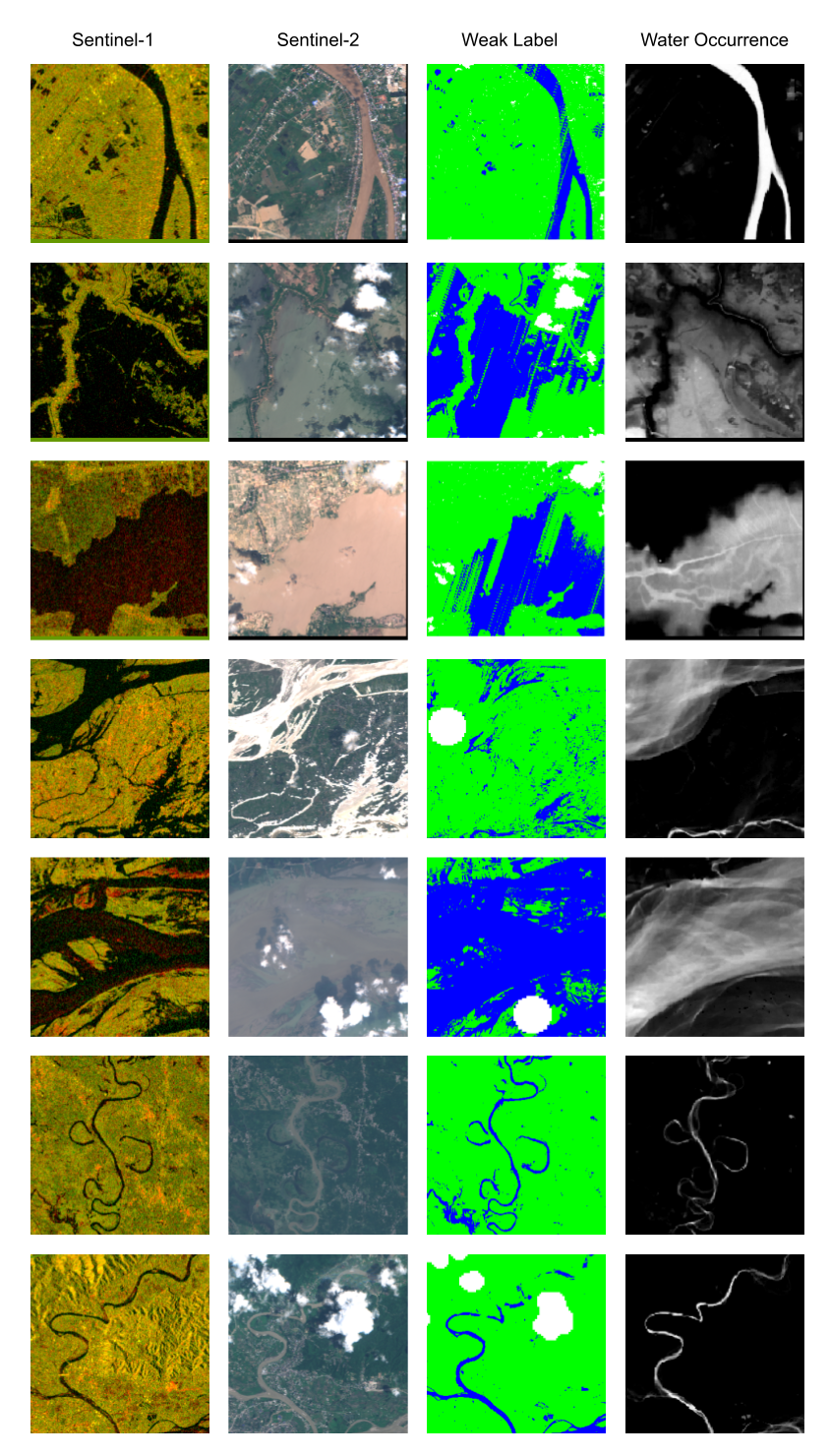}
\caption{Selectively sampled data points from Sen1Floods11 weak labelled data (first 3 rows) and Floods208 Dataset (last 4 rows). In the weak label the mapping is green:dry, blue:water and white:clouded/invalid pixels. These examples highlight the poor quality of the weak labels.}
\label{fig:s5}
\end{figure*}

The data points were extracted from 208 flooding events across Bangladesh, Brazil, Colombia, India and Peru. These regions are shown in Figure \ref{fig:s1} and were chosen according to the regions of interest for final deployment. We also extracted water occurrence map and a weak label for each image using the Normalized Difference Water Index (NDWI) band from the Sentinel-2 image. The extracted images had a resolution of 16m per pixel. Each image from a flooding event was then partitioned into multiple small tiles of size $320{\times}320$. The tiles at the edge of image were padded to fit the $320{\times}320$ tile size. Tiles having cloud percentage greater than $80\%$ were discarded. In total 23,260 valid tiles of size $320{\times}320$ were extracted from the whole process. Figure \ref{fig:s5} shows some selected data points from both the datasets.

\begin{table}[]
\resizebox{\textwidth}{!}{

\begin{tabular}{llll}
\toprule
                      & Sen1Floods11 ( hand label)                                                                                                                 & Sen1Floods11 ( weak label)                                                                                                                     & Floods208 dataset ( weak label)            \\
\midrule                      
No. train tiles      & 252                                                                                                                                        & 4385                                                                                                                                           & 23260                                     \\
No. validation tiles & 89                                                                                                                                         & -                                                                                                                                              & -                                         \\
No. test tiles       & 90                                                                                                                                         & -                                                                                                                                              & -                                         \\
Tile Size & $320\times320$ & $320\times320$ & $320\times320$ \\
Resolution & 16m & 16m & 16m \\
Sentinel-1 bands      & VV, VH                                                                                                                                  & VV, VH                                                                                                                                      & VV, VH                                 \\
Sentinel-2 bands      & B2, B3, B4, B8                                                                                                                             & -                                                                                                                                              & B2, B3, B4, B8                            \\
Sampling regions      & Bolivia, Colombia, Ghana, & Bolivia, Colombia, Ghana, & Bangladesh, Brazil, Peru,\\
 & India, Cambodia, Nigeria,  & India, Cambodia, Nigeria, & India, Columbia \\
 & Pakistan, Paraguay, USA, & Pakistan, Paraguay, USA, & \\
 & Sri-Lanka, Somalia & Sri-Lanka, Somalia & \\
\bottomrule
\end{tabular}}
\vspace{1em}
\caption{Summary of the key attributes of all the datasets used for training and evaluation.}
\label{tab:t1}
\end{table}

\subsection{Preprocessing}
The provided Sen1Floods11 dataset has Latitude / Longitude Projection and has a resolution of 10m per pixel. To match the projection and resolution of our Floods208 dataset, all the images are scaled to 16m per pixel input resolution and projected to the Universal Transverse Mercator (UTM) coordinate system. For Sentinel-1 image normalization, VV band was clipped to [-20, 0] and VH to [-30, 0] and then linearly scaled these values to the range [0,1]. For Sentinel-2 image, the 4 bands were clipped to [0, 3000] range and then linearly scaled them to [0, 1] range. Table \ref{tab:t1} summarises the final attributes of both the datasets.

\renewcommand{\thefootnote}{\fnsymbol{footnote}}
\section{Methods}
Our aim is to segment flooded pixels using Sentinel-1 SAR image as an input at inference time. Formally, let $X_{S1} \in R^{H\times W \times 2}$ be the SAR input space and let $Y \in R^{H \times W \times K}$ denote the pixel wise $K$ class one hot label in the output space ($K=2$ classes in our case: dry and floodded pixels). The paired Sentinel-2 images used in the training data are represented by $X_{S2} \in R^{H\times W \times 4}$.
The hand labelled training set is denoted by $D_l = \bigl\{X_{S1}^i, X_{S2}^i, Y^i\bigr\}_{i=1}^{N_l}$ and the larger weak labelled training set as $D_{wl} = \bigl\{X_{S1}^i, X_{S2}^i, \hat{Y}^i\bigr\}_{i=1}^{N_{wl}}$. Here $Y$ denotes a high quality label and $\hat{Y}$ denotes a noisy weak label. Our goal is to leverage both $D_l$ and $D_{wl}$ to train the segmentation network. The next section describes the supervised baseline, an approach to improve weak labels and the cross modal distillation framework.

\subsection{Supervised baseline}
\label{subsection:supervised}
We train two supervised models for comparison. The first model is trained only on hand labelled data $D_l$. The second model is trained only on the larger weak labelled dataset $D_{wl}$. The large size of the weak labeled data helps the network to generalize better (despite there model learning some of those label errors during training) ~\cite{rolnick2017deep}. We use Deeplab v3+~\cite{chen2018encoder} with an Xception 65 encoder~\cite{chollet2017xception} as the model architecture. Common regularization techniques like data augmentations (random crop with distortion, horizontal/vertical flips and colour jitter), dropout, weight decay and batch normalization are used to improve generalization. We also tried to fine tune the second model i.e the model trained on weak labelled data, using hand labelled data $D_l$. However, this additional training step did not improve the performance, so we decided not to include it in the baseline model.

\textbf{Edge weighted loss}

The network is trained to minimize the cross entropy loss. For every batch of images ( $B_{wl}$ ), the parameters $f$ of the network are updated by minimizing the cross entropy loss given by:

$$L = \frac{1}{B_{wl}} \sum l_{ce}(f(X_i), \hat{Y}_i, W_i)$$

\begin{figure*}[]
\centering
\includegraphics[width=11cm]{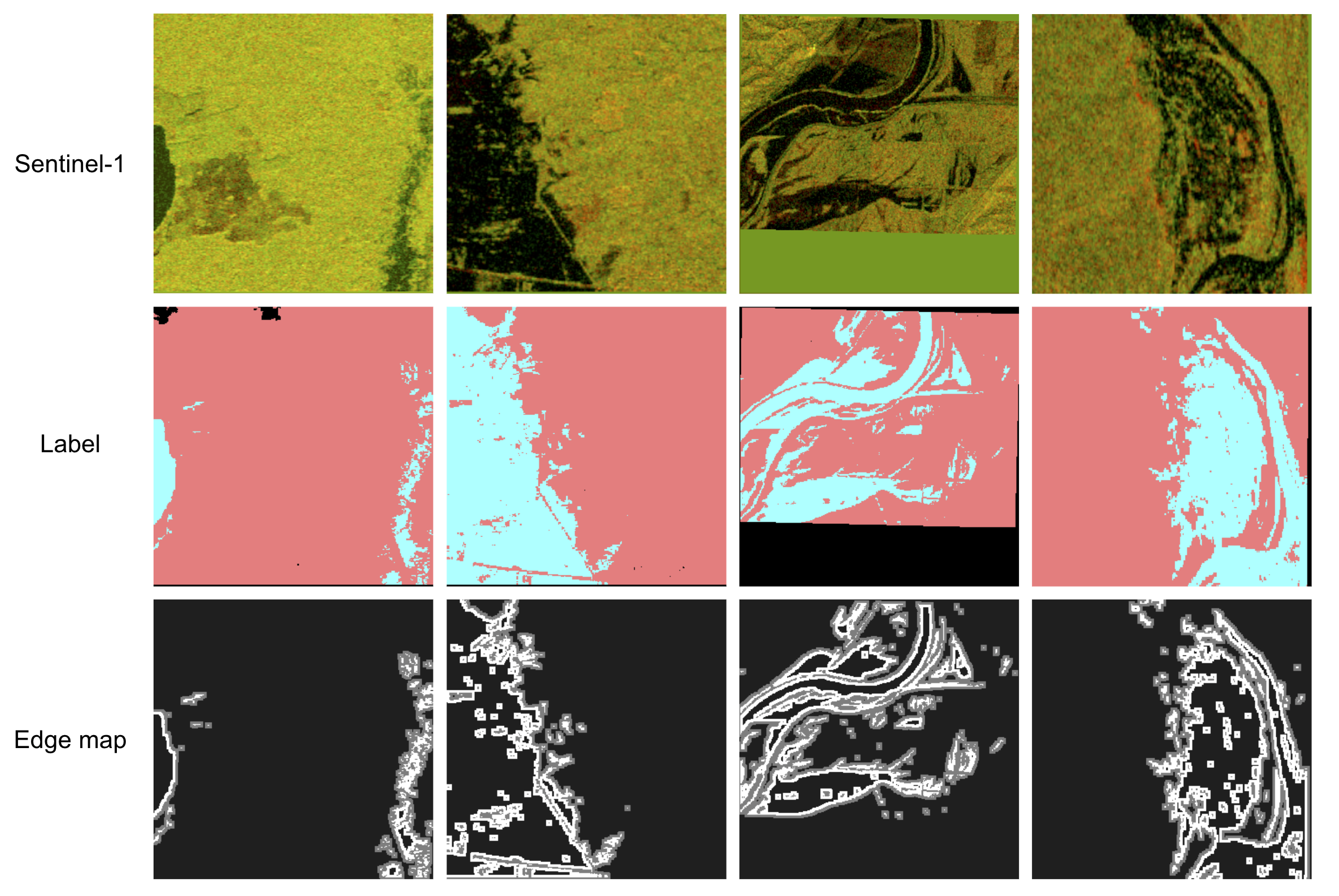}
\caption{Randomly selected Sentinel-1 images from the training split with their corresponding label and edge map. In the label, blue pixels denote water, peach pixels denote dry region and black pixels denotes the invalid pixels. The edge map shows the inner and outer edges in white and grey color respectively. }
\label{fig:s4}
\end{figure*}

Here $W_i$ represents the pixel wise weights for the $i^{th}$ image - $X_i$, applied to the cross entropy loss. We apply an edge based weighting that gives higher weights to the edges of the binary label. As suggested by~\cite{wu2016bridging}, from all the pixels in an image available for training, the pixels lying in the middle of an object can be easily discriminated by the model and are usually classified correctly. Hence, an edge weighted loss helps the network to focus on the harder to segment regions lying on the boundary during training. We compute two kinds of edges, inner and outer edges. Inner edges are obtained by subtracting the eroded labels from the original labels. Outer edges are obtained by subtracting the original labels from the dilated labels. All the other pixels are given a unit weight. The weights for inner and outer edges are decided by tuning these parameters during training. Figure \ref{fig:s4} shows these edges.

\subsection{Improving weak labels}
\label{subsec:weak_label_improv}

\begin{figure*}[]
\centering
\includegraphics[width=12cm]{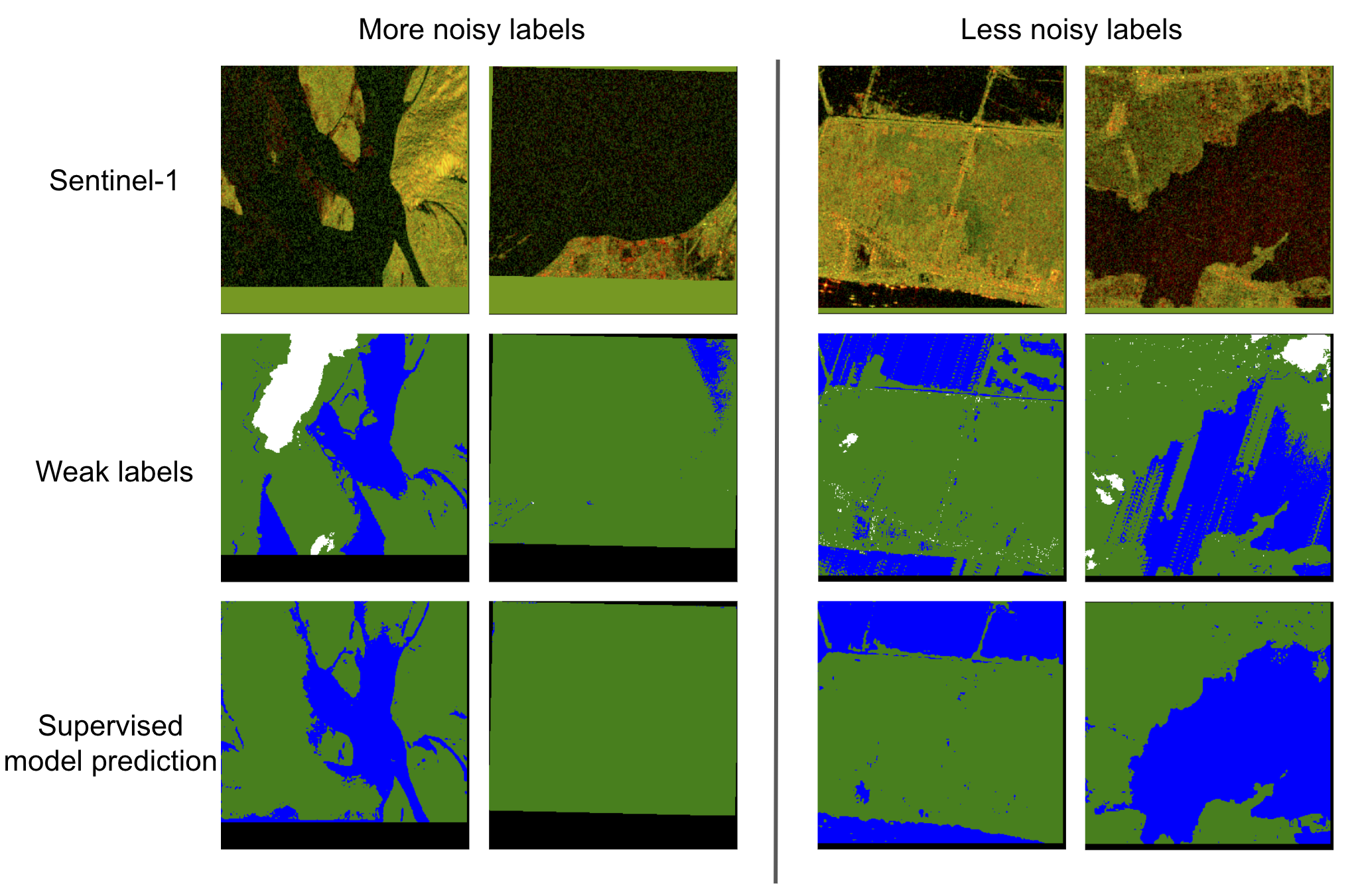}
\caption{Selected examples from the training split demonstrating the effect of amount of label noise in memorizing the label mistakes. Left: The model learns these mistakes when the label is of very poor quality and is missing most part of the river.Right:  The model can overcome these mistakes when there is less noise. In the label and prediction, green pixels are dry region, blue pixels are wet region, black are invalid region and white are clouded regions.}
\label{fig:s7}
\end{figure*}

Despite using regularization techniques, we notice that the model learns the label mistakes present in the training data. Figure \ref{fig:s7} demonstrates this on the training set images. Often such mistakes consist of cases where a complete or a major part of the river is missing. While the network is resilient to learning small label mistakes (such as slightly overflooded pixels surrounding the true boundary or random noise similar to cut out augmentation), it does tend to overfit to cases when the label mistakes are large. This is because from an optimization perspective, pixels in an extremely noisy label will dominate the loss, forcing the network to overfit on them in order to reduce the loss.

\begin{figure*}[]
\centering
\includegraphics[width=14cm]{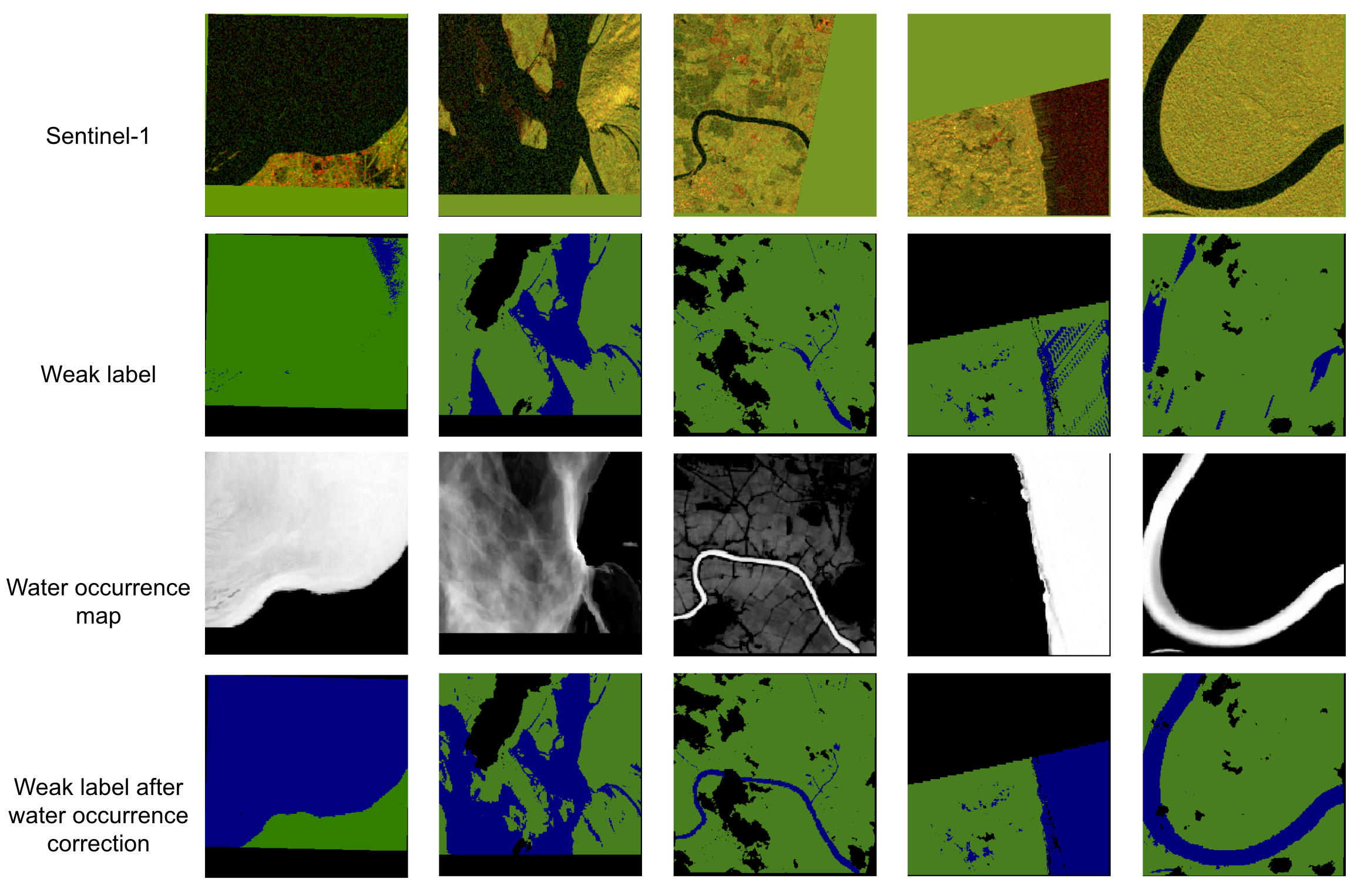}
\caption{Improving weak label from the training split using water occurrence map. In the weak label green pixels represent dry region, blue represents wet regions and black pixels mark the invalid pixels(out of bounds pixels)/clouded pixels in Sentinel-2 image.}
\label{fig:s8}
\end{figure*}

We aim to correct the labels having large parts of the river missing by using water occurrence map to get a rough estimate of such missing rivers. The label initially misses the river because they are formed by thresholding Sentinel-2 image and the label is sensitive to the value of the threshold. Since the water occurrence map~\cite{pekel2016high} is made by averaging long-term overall surface water occurrence, the permanent water pixels will have higher probability values and areas where water where sometimes occurs will have lower probability values. The weak label is improved by additionally marking the pixels having their water occurrence probability above a certain threshold, as wet in the label. This method will not be able to capture the seasonal/flooded or ephemeral water pixels, but these pixels mostly constitute a small majority of the label and as seen earlier, the model can learn to overcome such small mistakes in the labels. Figure \ref{fig:s8} shows some randomly selected tiles before and after the label improvement from the training set.

\subsection{Cross modal distillation}
\label{subsec:cross}
Though water occurrence map improved the weak labels in areas which are permanently water, it does not help with label mistakes in seasonal/flooded pixels. As a result, we can still get grossly incorrect weak labels in rare situations where flooded pixels constitute a majority portion of the water. To overcome this, we  explore a cross modal distillation framework that uses a network trained on a richer modality with a small set of hand labels, to generate more accurate labels on the training split. A teacher-student setup is used to transfer supervision between the two modalities. The teacher is trained on stacked Sentinel-1 and Sentinel-2 images using accurate labels from $D_l$, and is used to supervise a Sentinel-1 only student model on the unlabelled images from $D_{wl}$. The advantage of this method over binary weak labels (used in Section \ref{subsection:supervised}), is that the soft labels predicted by the teacher capture uncertainty better compared to the binary labels ~\cite{ba2014deep, hinton2015distilling}. The soft targets are also outputs of a network trained on hand labeled data and are more likely to be correct than weak labels. Compared to self distillation, cross modal distillation enables us to provide more accurate supervision by transferring information from a richer knowledge modality. Figure \ref{fig:s} summarizes the training setup used in our work. Both the teacher and the student have identical architecture backbones. Let $f_t$ and $f_s$ represent the teacher and student network function, respectively. The training is done in two stages as described below:

\begin{figure*}[]
\centering
\includegraphics[width=13cm]{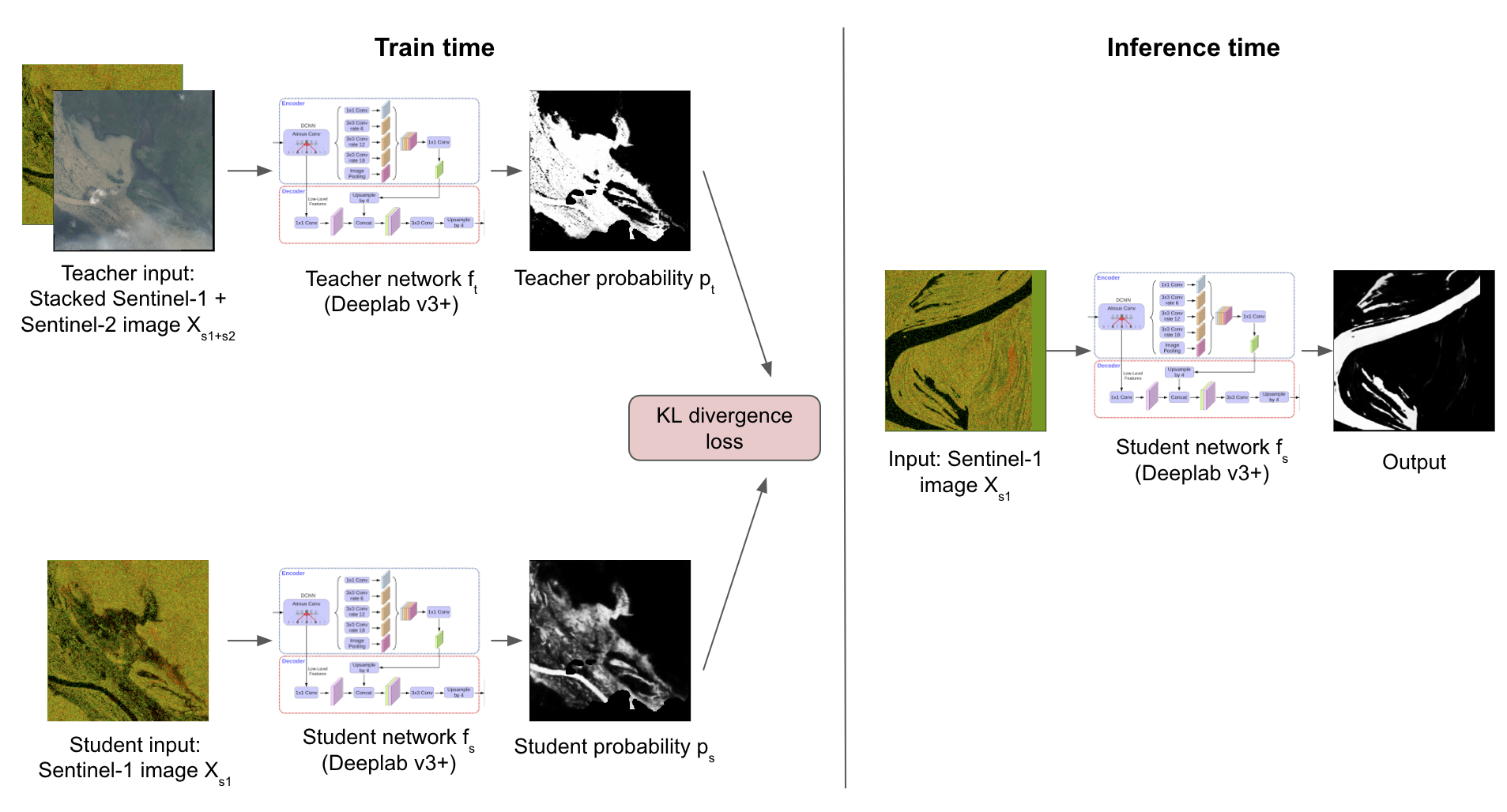}
\caption{Overview of our cross modal distillation framework. In the training stage, a teacher model using Sentinel-1 and Sentinel-2 images is used to train a student using only the Sentinel-1 image. At inference time, only the student is used to make predictions.}
\label{fig:s}
\end{figure*}

\textbf{Stage 1: Training the teacher network} 
Let $X_{S1+S2}^i$ denote the $i^{th}$ stacked Sentinel-1 and Sentinel-2 image. $X_{S1+S2}^i \in D_l$ is used as input to the teacher network. The teacher is trained in the same manner as the supervised baseline described in Section \ref{subsection:supervised}. The training set is small but contains data from geographic locations spanning 6 different continents. This helps the teacher generalize well to different geographies in the unlabelled data seen during the next stage of training. 

\textbf{Stage 2: Training the student network}
The teacher weights from Stage 1 are kept frozen in this stage. We use paired Sentinel-1 and Sentinel-2 images from Sen1Floods11 data(i.e. both hand labelled and weak labelled) and Floods208 weak labelled data as the unlabelled data to train the student network. The data in each batch is sampled equally from both the data sources to ensure equal weighting for the datasets.
The stacked Sentinel-1 and Sentinel-2 image $X_{S1+S2}^i$ is passed through the teacher to obtain the probabilities $p_t = \sigma (f_t (X_{S1+S2}^i))$ and the augmented paired Sentinel-1 image $\tilde{X}_{S1}^i = Aug(X_{S1}^i)$ is passed through the student to get the student probabilities $p_s = \sigma (f_s (\tilde{X}_{S1}^i))$. Here $\sigma$ refers to the sigmoid function used to convert the model's output into a probability score. KL divergence loss ($L_{KD}$) is then minimized for $K=2$ classes to update the student weights: 
$$ L_{KD} = -\sum_{i=1}^K p_t \log{p_s} $$

\section{Experiments and Results}
\subsection{Training details}

For all the models, we use Deeplab v3+ model~\cite{chen2018encoder} with Xception 65~\cite{chollet2017xception} as the backbone encoder. The skip connection from the encoder features to the decoder is applied at stride 4 and 2. We use a batch size of 64 with input image shape of $(321, 321, C)$ (here $C = 2$ for Sentinel-1 images and $C = 4$ for Sentinel-2 images). For optimization, Momentum optimizer is used with momentum set to 0.9. The learning rate is decayed with a polynomial schedule from initial value to zero with a power of 0.9. The models are trained for 30k steps. A learning rate and weight decay grid search hyperparameter tuning is done by choosing a learning rate from \{0.3, 0.1, 0.003, 0.001\} and weight decay from \{1e-3, 1e-4, 1e-5, 1e-6\}. For edge weighted cross entropy loss, the inner and outer edge weight is set to 10 and 5 respectively. A threshold of value 0.5 is used to create a binary mask of permanent water pixels from the water occurrence map. All the hyperparameter tuning and best model checkpoint selection is then done on the validation split. After the best checkpoint selection, the model is frozen and all the results are reported on the test split.

\subsection{Evaluation Metric}
We use pixel-wise intersection over union (IoU) of the water class to validate our model performance. It is defined as follows:

$$ IoU = \frac{\sum_{i=1}^{N} {TP}_i}{\sum_{i=1}^{N} ({TP}_i + {FP}_i + {FN}_i )}  $$

In the above formula, $i$ is the iterator over all the images and $N$ represents the total number of images. For the $i^{th}$ image, ${TP}_i$, ${FP}_i$ and ${FN}_i$ denotes the number of true positives, false positives and false negatives of water class respectively. 

\subsection{Results}
\textbf{Comparison with our baselines}

Table \ref{tab:1} shows the result of the supervised baseline model trained on the Sen1Floods11 hand labelled data. As expected, the model using Sentinel-2 as input image is much higher than the models using Sentinel-1 image because Sentinel-2 image is a richer modality. However, Sentinel-2 image is not suitable for inference because of cloud cover issues mentioned in Section \ref{subsection:input}. Hence, even though we can't use the model using stacked Sentinel-1 and Sentinel-2 image during inference, we can opportunistically utilize this as a richer set of features to train the teacher model for cross modal distillation as discussed in Section \ref{subsec:cross}.

\begin{table}[]
\centering
\begin{tabular}{lll}
\toprule
\textbf{Input image}             & \textbf{Bands}                                                           & \textbf{IoU}  \\
\midrule
Sentinel-1              & VV, VH                                                          & $67.63 \pm 0.45 $ \\
\noalign{\vskip 2mm} 
Sentinel-2              & B2, B3, B4, B8                                                  & $79.02 \pm 1.05$ \\
\noalign{\vskip 2mm} 
Sentinel-1 + Sentinel 2 & VV, VH & $79.25 \pm 1.07$\\
 & B2, B3, B4, B8 & \\
\bottomrule
\end{tabular}
\vspace{1.5mm}
\caption{Test split results of our model trained on Sen1Floods11 hand labelled data at 16m resolution. The numbers show the aggregated mean and standard deviation of IoU from 5 runs.}
\label{tab:1}
\end{table}

\begin{table}[]
\centering
\setlength\belowcaptionskip{-5pt}
\begin{tabular}{ll}
\toprule
\textbf{Method}                                                                                 & \textbf{IoU}  \\
\hline
Hand labelled supervised                                                                &  $67.63 \pm 0.45$ \\
Weak labelled supervised: Sen1Floods11 weak       & $67.76 \pm 2.41$ \\
Weak labelled supervised: Sen1Floods11 + Floods208 weak & $68.94 \pm 1.11$ \\
Weak labelled supervised: Floods208 weak   & $67.53 \pm 0.19$ \\
Improved weak labelled supervised: Sen1Floods11 + Floods208 weak & $70.64 \pm 0.91$ \\
Cross modal distillation          &                                                    \boldmath{$71.86\pm0.91$}\\
\bottomrule
\end{tabular}
\vspace{1.5mm}
\caption{Result of our Sentinel-1 supervised baseline models, improved weak label supervised model and our cross modal distillation framework on Sen1Floods11 handlabel test split at 16m resolution. The numbers show the aggregated mean and standard deviation of IoU from 5 runs.}\label{tab:2}
\end{table}

\begin{figure*}[h!]
 \centering
\includegraphics[width=12cm]{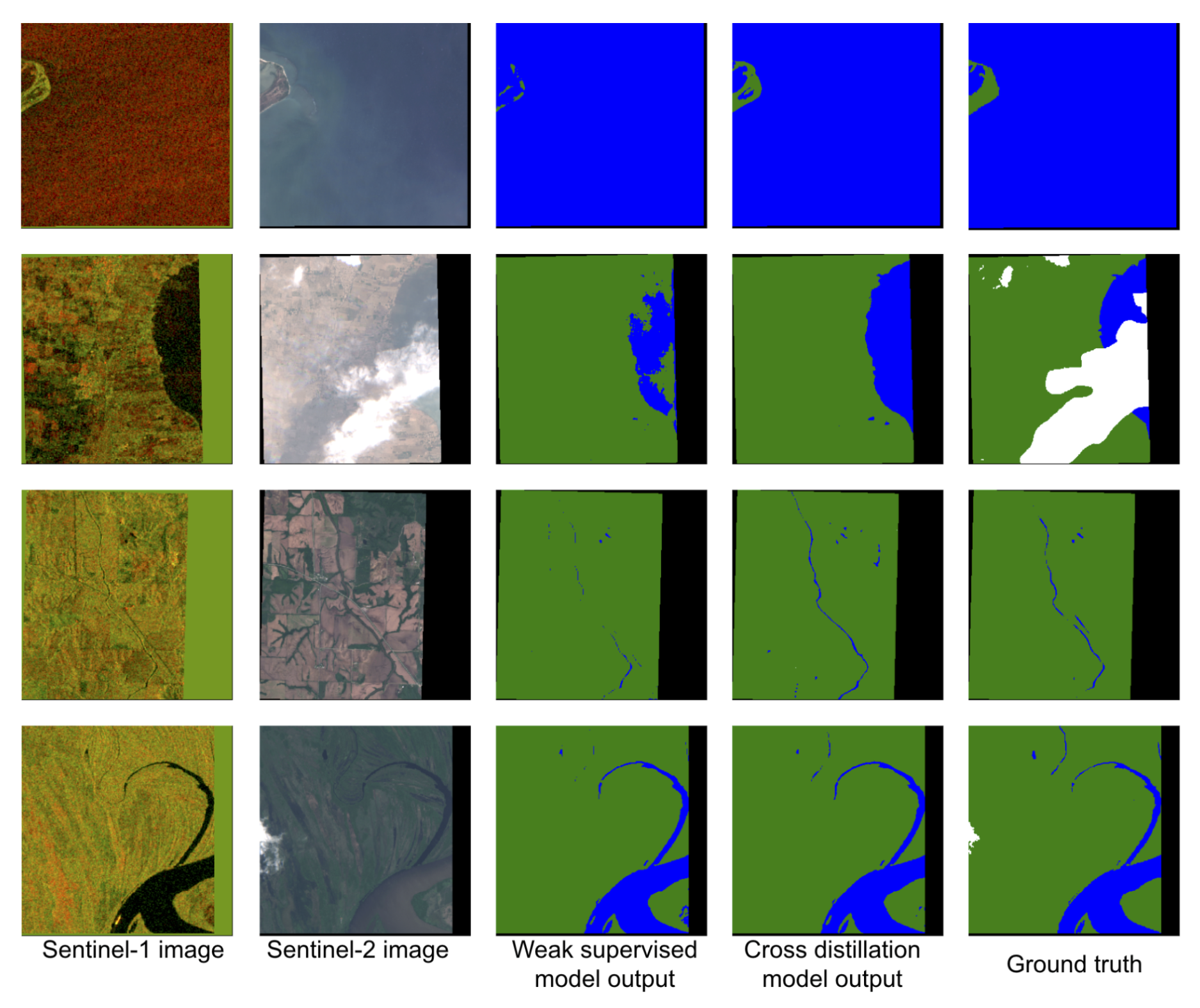}
\caption{Model inference visualisation on Sen1Floods11 hand labelled test split on selected images. In the output predictions dry pixels are shown in green, water pixels in blue and invalid pixels in black. The Sentinel-2 image is not passed as input to the model and only shown for visualisation purpose. The ground truth is hand labelled on the Sentinel-2 image and can contain clouds which are masked in white color. It can be seen that cross modal distillation produces sharper and more accurate results. Weak labelled supervised baseline on the other hand sometimes misses big parts of river due to mistakes learnt from the training data.}
\label{fig:s9}
\end{figure*}

Table \ref{tab:2} shows a quantitative comparison of the supervised baseline method with model trained after improving the weak labels and the cross distillation model. In Table \ref{tab:2} first two rows, it can be seen that a supervised model trained on Sentinel-2 weak label can match the performance of a supervised model trained on small set of hand labelled data. This empirically verifies the claim that a large weak labelled dataset can act as a quick substitute for a small amount of costly hand label annotations. Including Floods208 weak labelled data to Sen1Floods11 weak labelled data, further led to an increase in the model performance by $1.18\%$ from Sen1Floods11 weak label supervised baseline. This shows that there were still more gains to be had by increasing the weak label dataset size. Even though Floods208 weak labelled dataset by itself is also a very large dataset, the model trained on only Floods208 does not perform good. This is because Floods208 data is localised to a few specific locations while Sentinel-2 weak labelled split has data from 6 continents across the world. This emphasizes the importance of sampling training data strategically, such that the locations are more global and includes various topographies. We can also see that improving weak labels using water occurrence map as described in Section \ref{subsec:weak_label_improv} increases the model IoU by $1.7\%$ compared to its weak supervised counterpart and by $2.88\%$ compared to the Sen1Floods11 weak label baseline. This emphasises the importance of quality of training labels in producing more accurate models. Our cross distillation model further produces improves the quality of training label using a teacher model and performs better than all the other models. It exceeds Sen1Floods11 weak label baseline by $4.1\%$ IoU and improved weak label supervised model by $1.22\%$ IoU. Figure \ref{fig:s9} shows a qualitative comparison of the cross distillation model with the baseline model.

\textbf{Comparison with other methods}

\begin{table}[H]
\centering
\setlength\belowcaptionskip{-5pt}
\begin{tabular}{ll}
\toprule
\textbf{Method}                                                                                 & \textbf{IoU}  \\
\hline
Sen1Floods11 Otsu thresholding \cite{bonafilia2020sen1floods11} & $54.58$ \\
Sen1Floods11 Sentinel-2 Weak label model \cite{bonafilia2020sen1floods11} & $66.21$ \\
BASNet \cite{bai2021enhancement}  & $53.90$ \\
AN-34 \cite{helleis2022sentinel} & $49.70$ \\
S-1FS \cite{helleis2022sentinel} & $54.90$ \\
Cross modal distillation (Ours)          &                                                    \boldmath{$72.74$}\\
\bottomrule
\end{tabular}
\vspace{1.5mm}
\caption{Performance comparison of our cross modal distillation model with other methods on all water hand labels from Sen1Floods11 test set at 10m resolution. }\label{tab:c1}
\end{table}

Benchmark comparisons of IoU on Sen1Floods11 test split are provided in Table \ref{tab:c1}. Note that even though our method is trained at input images with resolution of 16m, here we do an evaluation at 10m Sentinel-1 images and hand labels for a fair comparison with other methods. To get an inference output at a resolution of 10m, we first downsample the original image of shape $512\times512$ to a resolution of 16m i.e. $320\times320$, feed the downsampled image to the model and then  upsample the probabilities to their original resolution.

From Table \ref{tab:c1} we can see that our method outperform Sen1Floods11 ~\cite{bonafilia2020sen1floods11} weak label baseline by a absolute margin of 6.53\% IoU.

\vspace{-1em}

\section{Sensitivity Analysis}
\subsection{Effect of decoder stride}

In the Deeplabv3+~\cite{chen2018encoder} architecture, an input image is first passed to an encoder to generate semantically meaningful features at different scales.  The final encoder output feature has an output stride of 16, i.e. the input image is down-sampled by a factor of 16. This feature is then passed through ASPP (Atrous Spatial Pyramid Pooling) layers and the decoder to produce the output logits. The decoder processes the encoder features at a list of multiple strides - a hyper-parameter that can be tuned to refine the output segmentation masks. At each stride, the low level features from the encoder having the same spatial dimensions, are concatenated with the decoder features to process them further. The smallest stride in the list refers to the final ratio of the original input image size with the final logits size. The final logits are then bi-linearly upsampled to the required input size. Table \ref{tab:a1} shows the effect of the decoder stride on the validation split results. It can be seen that as we concatenate finer resolutions features from the encoder with the decoder features, the results improve. However, the rate of improvement declines. Considering practical considerations related to the number of model parameters, we decided to go with decoder stride (4, 2). 

\begin{table}[H]
\centering
\begin{tabular}{ll}
\toprule
\textbf{Method}        & \textbf{IoU} \\
\midrule
Decoder stride 4       & $66.63 \pm 1.04$          \\
Decoder stride 4, 2    & $67.76 \pm 1.41$            \\
Decoder stride 4, 2, 1 & $67.79 \pm 1.67$      \\
\bottomrule
\end{tabular}
\vspace{1.5mm}
\caption{Validation split results for decoder stride comparison on Sen1Floods11 hand labeled split. The numbers show the aggregated mean and standard deviation of IoU from 5 runs.}
\label{tab:a1}
\end{table}

\subsection{Effect of loss function}
We also show the comparison of different loss function in Table \ref{tab:a2} on the validation split. As can be seen from the table that vanilla cross entropy loss performs poorly. On the other hand, cross entropy with edge weighting improves vanilla cross entropy loss by $9.68\%$ IoU. Also Tversky focal loss outperforms edge weighted Cross Entropy loss by a margin of $1.91\%$ IoU. However, we observed that the water class probability outputs with Tversky focal loss were not calibrated. A neural network is said to be calibrated if its output confidence is equal to the probability of it being correct. Compared to Tversky loss that classifies both correct and incorrect water pixels with high confidence, Cross Entropy loss produces a more calibrated output. Because a calibrated output helps in getting a better estimate of the uncertainty of the model prediction, we decided to go with edge weighted Cross Entropy loss.

\begin{table}[H]
\centering
\begin{tabular}{ll}
\toprule
\textbf{Method}                & \textbf{IoU} \\
\midrule
Cross Entropy                  & $58.08 \pm 2.85$             \\
Cross Entropy + Edge weighting & $67.76 \pm 1.41$            \\
Tversky focal loss             & $69.67 \pm 0.84$            \\
\bottomrule
\end{tabular}
\vspace{1.5mm}
\caption{Validation split results for loss comparison on Sen1Floods11 hand labeled split. The numbers show the aggregated mean and standard deviation of IoU from 5 runs.}
\label{tab:a2}
\end{table}

\section{Conclusion}
We proposed a simple cross modal distillation framework to effectively leverage large amounts of unlabelled and paired satellite data and a limited amount of high quality hand labelled data. We distill knowledge from a teacher trained on the hand labelled images using the more informative modality as input. This helped us generate more accurate labels for the student network as compared to weak labels created by a simple thresholding technique. The student network trained this way outperforms both the supervised hand label and weak label baselines. A promising avenue for future research would be to include temporal imagery to improve performance.

\section{Acknowledgements} 
We would like to thank Yotam Gigi and John Platt for helping us review the paper and providing invaluable suggestions throughout the whole process.

\bibliographystyle{plain}
\bibliography{references.bib}
\newpage




\end{document}